\documentclass[letterpaper, 10 pt, conference]{ieeeconf}
             
\IEEEoverridecommandlockouts

\usepackage{siunitx}
\usepackage{xcolor}

\newcommand{\gtsensedlocations}{\mathcal{X}}
\newcommand{\obstacle}{\mathcal{O}}
\newcommand{\obstacles}{\mathcal{O}}
\newcommand{\lightintensity}{L}

\newcommand{\argmin}{\arg\!\min}

\newcommand{\mW}{\mathcal{W}}
\newcommand{\vx}{\mathbf{x}}
\newcommand{\vp}{\mathbf{p}}
\newcommand{\vs}{\mathbf{s}}
\newcommand{\mR}{\mathcal{R}}

\newcommand{\reals}{\mathbb{R}}

\usepackage[backend=biber,style=ieee,natbib=true,doi=false,isbn=false,url=false,mincitenames=1]{biblatex} 
\usepackage{amsmath,amssymb,amsfonts}
\usepackage{algorithmic}
\usepackage{graphicx}
\usepackage{textcomp}
\usepackage{color}
\usepackage[font=footnotesize]{caption}
\usepackage[font=footnotesize]{subcaption}
\usepackage{cleveref}
\crefname{algocf}{alg.}{algs.}
\Crefname{algocf}{Algorithm}{Algorithms}
\usepackage{amsmath}
\usepackage{amssymb}
\usepackage{fixmath}
\usepackage{siunitx}
\usepackage[noend]{algorithm2e}
\SetAlFnt{\footnotesize}
 \RestyleAlgo{ruled}
\def\BibTeX{{\rm B\kern-.05em{\sc i\kern-.025em b}\kern-.08em
    T\kern-.1667em\lower.7ex\hbox{E}\kern-.125emX}}
\PassOptionsToPackage{hyphens}{url}
\begin{document}
\title{
\LARGE \bf Active Signal Emitter Placement In Complex Environments
}

%TODO workshop title? 
% the environment is not really unkonwn, maybe cluttered
% signal emission is not really correct
% finish abstract
% add more to the electomagnetic emitter portion (integrate lillys papers)
%de-plagiarize common background
%add spaces that formulation symbols live in

\author{Christopher E. Denniston, Bask{\i}n \c{S}enba\c{s}lar, and Gaurav S. Sukhatme%
\thanks{Christopher E. Denniston  is with Parallel Systems, Bask{\i}n \c{S}enba\c{s}lar is with NVIDIA, and Gaurav S. Sukhatme is with the Robotic Embedded Systems Lab at the University of Southern California. This paper describes a work when Denniston and \c{S}enba\c{s}lar were PhD students at USC and is not associated with Parallel Systems or NVIDIA. Sukhatme holds concurrent appointments as a Professor at USC and as an Amazon Scholar. This paper describes work not associated with Amazon. Corresponding Author: cedenniston@gmail.com.}
}

\maketitle

\begin{abstract}
Placement of electromagnetic signal emitting devices, such as light sources, has important usage in for signal coverage tasks.
Automatic placement of these devices is challenging because of the complex interaction of the signal and environment due to reflection, refraction and scattering.
In this work, we iteratively improve the placement of these devices by interleaving device placement and sensing actions, correcting errors in the model of the signal propagation.
To this end, we propose a novel factor-graph based belief model which combines the measurements taken by the robot and an analytical light propagation model.
This model allows accurately modelling the uncertainty of the light propagation with respect to the obstacles, which greatly improves the informative path planning routine. 
Additionally, we propose a method for determining when to re-plan the emitter placements to balance a trade-off between information about a specific configuration and frequent updating of the configuration.
This method incorporates the uncertainty from belief model to adaptively determine when re-configuration is needed.
We find that our system has a 9.8\% median error reduction compared to a baseline system in simulations in the most difficult environment.
We also run on-robot tests and determine that our system performs favorably compared to the baseline.
\end{abstract}

\section{Introduction}

% Outline:

% \begin{itemize}
%     \item Introduce problem
%     \item motivate problem
%     \begin{itemize}
%         \item 
%     \end{itemize}
%     \item Explain ambient lighting and how analytical model can be wrong
% \end{itemize}

Electromagnetic wave emitting devices such as ultrawideband beacons (UWB), Wi-Fi access points, radios, and light sources are commonly used for localization, communication, and illumination. 
The coverage provided by these devices is highly dependent on the exact locations of them in complex obstacle rich environments.
Electromagnetic waves interact with obstacles in the environment by reflection, refraction, and scattering depending on surface qualities and the nature of the wave.
Often, models of electronmagnetic wave propagation fail to exactly predict how the wave will interact with the environment due to unmodeled effects, imprecise parameter specification, and previously unknown emitting sources.\looseness=-1

\begin{figure}[t]
    \centering
    \begin{subfigure}{.49\columnwidth}
        \includegraphics[clip=true, trim={1.5cm 1.5cm 1.5cm 1.5cm},width=\textwidth]{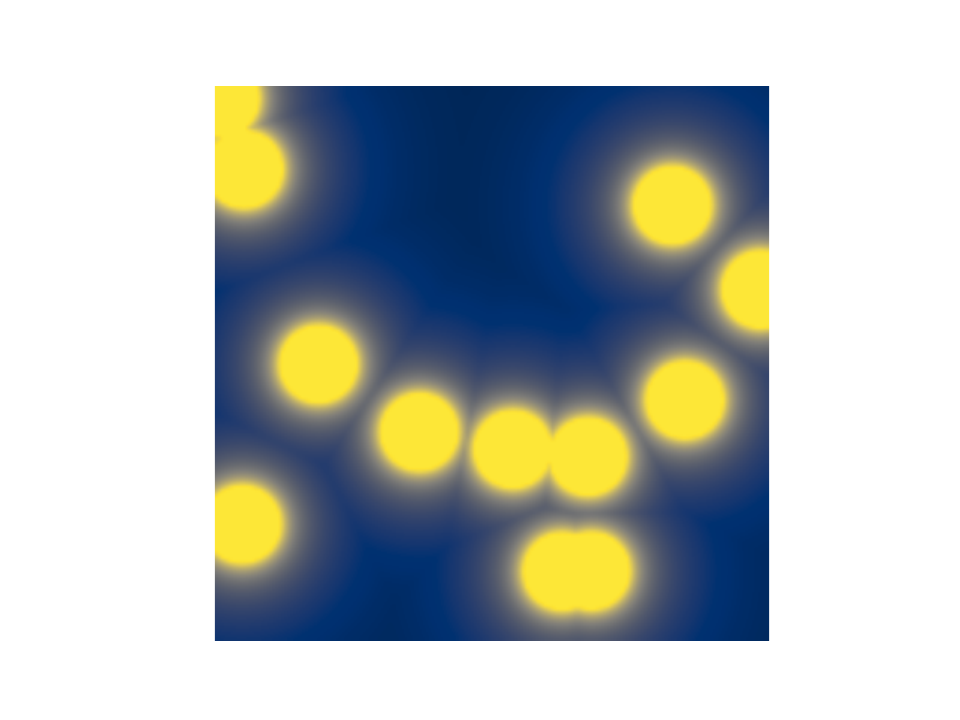}
        \caption{Desired Lighting}\label{fig:desired}
        \vspace{0.05in}
    \end{subfigure}
    \begin{subfigure}{.49\columnwidth}
        \includegraphics[clip=true, trim={1.5cm 1.5cm 1.5cm 1.5cm},width=\textwidth]{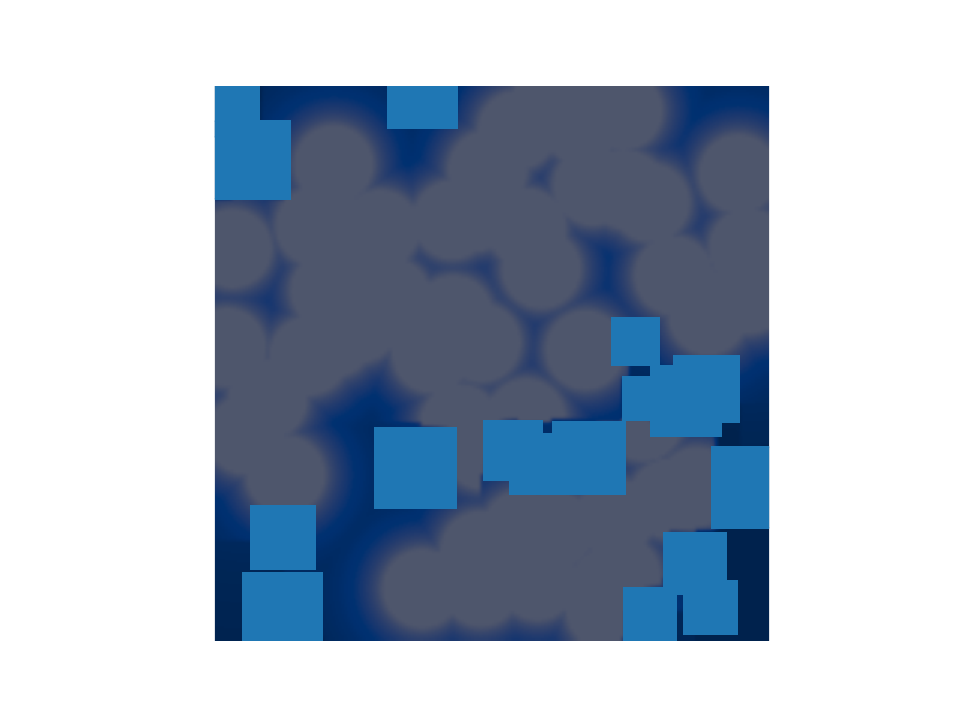}
        \caption{Unknown Lighting}\label{fig:unknown}
        \vspace{0.05in}
    \end{subfigure}
    \begin{subfigure}{.49\columnwidth}
        \includegraphics[clip=true, trim={1.5cm 1.5cm 1.5cm 1.5cm},width=\textwidth]{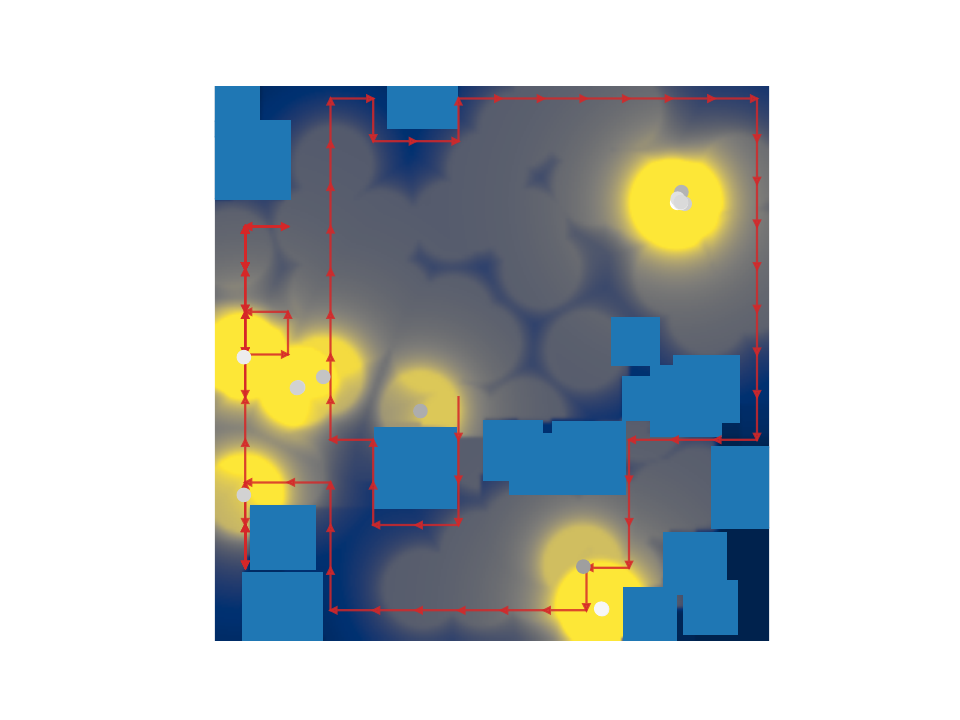}
        \caption{Proposed System Result}\label{fig:proposed}
    \end{subfigure}
    \begin{subfigure}{.49\columnwidth}
        \includegraphics[clip=true, trim={1.5cm 1.5cm 1.5cm 1.5cm},width=\textwidth]{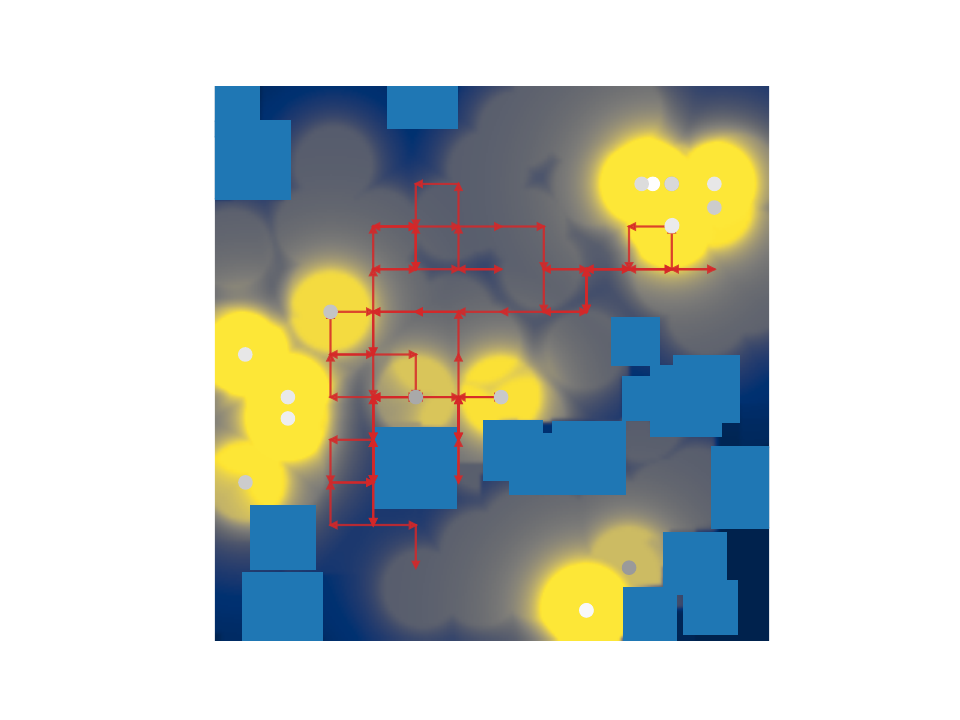}
        \caption{Baseline Result}\label{fig:baseline}
    \end{subfigure}
    \begin{subfigure}{\columnwidth}
    \vspace{0.05in}
           \includegraphics[width=\columnwidth]{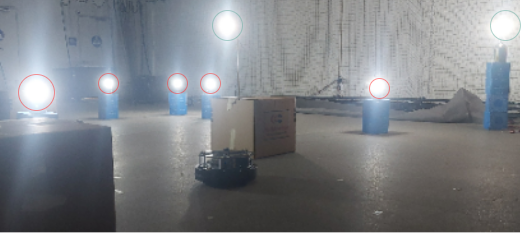}
           \caption{Field Trial}\label{fig:field_hero}
    \end{subfigure}

    \caption{\textbf{Overview:} a) The desired lighting intensity, chosen by randomly placing lights, b) the lighting intensity in the environment which is unknown to the robot a priori, as well as the obstacles, c) the final light source configuration, lighting intensity and robot path produced by the proposed system, d) 
    the final light source configuration,  lighting intensity and robot path produced by the baseline system, e) example field trial. Lights circled in red are deployed by the robot, while lights circled in green are unknown to the robot.
    }\label{fig:hero_setup}
    \vspace{-0.3in}
\end{figure}

In this work, we investigate the active signal emitter placement problem, focusing specifically on automatic deployment of light sources in obstacle-rich environments. 
We focus on light source placement due the visible interaction of light with obstacles, which allows us to qualitatively compare different methods in the real world, as well as  the importance of the task itself.
Actively placing lights can be used in applications ranging from search and rescue, where some locations have higher prioritization than others, to cinematic lighting in movie sets, as well as complex exploration scenarios.

Autonomously deploying emitters is a challenging task due to previously mentioned model prediction errors.
We propose a formulation where imprecise signal models are improved using in-situ signal strength measurements from sensors mounted on a robot, which are subsequently used for automatic deployment of emitters to match a user-specified desired signal strengths at specific locations. 
To plan to sense the environment, we use an informative path planning setup, with a novel factor-graph based environment representation which allows combining an analytical signal model with measurements taken from the environment to correct for errors in the analytical model.
The robot can only measure the combined signal strength of the placed emitters, their unmodeled effects, and the unknown background signal, making disambiguating these signals difficult.
As the robot senses the environment, it must decide when the appropriate time to re-configure the emitter placement is.
This causes the robot to trade-off between re-planning for emitter placement and collecting measurements.
Re-planning increases the accuracy of the resulting signal strength in the environment given robot's current information about the environment, but causes the robot to take fewer measurements between emitter replacements.
To this end, we propose a novel signal strength condition, i.e., \emph{the placement trigger}, which estimates the likelihood of observing the desired signal strength given the model's current information.
The robot iteratively moves to a new position, senses the signal strength at its current position, and decides whether it should re-plan emitter placement based on the placement trigger.
% Re-placing emitters allows the robot to account for imprecise sensing, unknown background signal, and unmodeled signal effects. 

% \begin{figure}[t]
%     \centering
%    \includegraphics[width=\columnwidth]{figures/light_hero.png}
%    \caption{\textbf{Field Trial Setup:} Example field trial. Lights circled in red are deployed by the robot, while lights circled in green are unknown to the robot. The environment also features uncontrolled light leakage due to windows and unsealed doors. The environment has many un-modeled light effects, such as scattering, as well as the white walls which enclose the area.}
% \end{figure}

The concrete setup for light source placement that we use for this problem can be seen in \cref{fig:hero_setup} where the baseline and proposed system in this work are compared. 
In~\cref{fig:desired}, the desired lighting is specified by the user.\footnote{In this work, the desired light placement is specified by placing lights in an obstacle-free environment.}
\Cref{fig:unknown} shows the unknown lighting, which is present at a lower intensity than the desired lighting, as well as the obstacles in the environment. 
\Cref{fig:proposed} and~\cref{fig:baseline} show example trajectories taken by the robot while sensing, as well as the final lighting placement in the environment for each system.

The contributions of our work are as follows:
\begin{itemize}
    \item We propose a novel factor-graph based formulation for signal emitter placement problem, which allows improving the accuracy of the analytical signal model using sensor data.
    \item We introduce a novel signal uncertainty condition, i.e. the reconfiguration trigger, which decides between data collection and re-planning emitter placement.
    \item We evaluate our method in simulations using visible light based experiments and show that it ourperforms a baseline method.
    We show our method's applicability to the real-world using physical robot experiments.
\end{itemize}

\section{Background / Related Work}
\textbf{Electromagnetic emitter placement}
In this work, we aim to optimally place light sources using in-situ measurements from a robot.
Other electromagnetic signals have been considered, with specific propogation properties.
The placement of UWB anchors have been extensively studied due to the effectiveness in localizing robots in GPS-denied environments.
UWB anchors placement has been optimized with genetic algorithms and a Gaussian process based residual model~\cite{pan2022uwb}.
It has also been noted that the exact placement of these beacons greatly affects their performance and it is difficult to analytically determine their range measurements at a given location~\cite{pan2022uwb,puspitasari2015wifi}.
Wi-Fi access point optimization has also been considered using line of site and non-line of site models, using non-convex optimization~\cite{puspitasari2015wifi}.
Base station placement for cellular mobile communication systems is investigated, in which global coverage~\cite{gamst1986coverage} or demand coverage~\cite{tutschku1998cellular} are common objectives.
Mesh node placement for wireless mesh networks with the objectives of coverage and connectivity is considered in~\cite{franklin2007mesh}, where mesh node locations are selected sequentially by greedily optimizing a performance metric.
Autonomously deploying robotic mesh network nodes in disaster scenarios using a two tiered connectivity, in which the connectivity between survivors are established via short-range Bluetooth Low Energy (BLE) and the connectivity between robots is established with long-range very high frequency (VHF) links is proposed~\cite{ferranti2022hironet}.
Distributed self-deployment of acoustic underwater sensors has been investigated, which can only move vertically, guaranteeing connectivity and increasing coverage using locally computable movement rules~\cite{akkaya2009underwater}.

%TODO make this not plagiarized and specific 
\textbf{Informative path planning} (IPP) is a process in which a robot takes measurements of a concentration to maximize an information metric.
Informative path planning can be used to minimize the uncertainty in the model, such as through an entropy metric~\cite{g2005}.
Previously, this formulation has been used in finding high concentrations of chlorophyll for studying algae blooms~\cite{mccammon_ocean_2021,fossum_information-driven_nodate}, or finding the quantiles of a chlorophyll distribution~\cite{denniston_ral}.

\textbf{Gaussian Processes} are used to model the environment from measurements with uncertainty quantification which are widely used for IPP~\cite{marchant_sequential_2014,denniston_ral,kemna_voronoi,online_radio,denniston_icra_2021}.
They approximate an unknown function from its known outputs by computing the similarity between points from a kernel function, $k$~\cite{Rasmussen2006}.
% Function values $y_*$ at any input location $\mathbf{x}_*$ are approximated by a Gaussian distribution:
%  $\sensedvalues_* | \sensedvalues \sim \mathcal{N} \left(K_* K^{-1} y,~ K_{**} - K_* K^{-1} K^T_* \right) $
% where
%  $\sensedvalues$ is training output,
%  $\sensedvalues_*$ is test output.
% % The $K$ are the kernel matrices over specific sets of data, such that
% %  $K$ is $k(\sensedlocations,\sensedlocations),$
% %  $K_*$ is $k(\sensedlocations,\sensedlocations_*),$
% %  $K_{**}$ is $k(\sensedlocations_*,\sensedlocations_*).$
% The kernel matrices $K$, $K_*$, and $K_{**}$ are computed by evaluating the kernel on $\sensedlocations$, the training input, and
%  $\sensedlocations_*$, the test input, such that $K^{i,j} = k(x^i,x^j)$, $K_*^{i,j} = k(x^i, x_{*}^j)$, and $K_{**}^{i,j} = k(\sensedlocation_{*}^i, \sensedlocation_{*}^j)$.

Gaussian processes typically do not have a way to represent obstacles as the kernel function only relies only on the distance between the locations in the model.
Gaussian processes tend to scale cubically  in the time required for inference due to the need to invert a matrix with entries for each collected measurement, and scale linearly in the time required when adding new measurements as the kernel function needs to be computed with the new measurement and all previous measurements~\cite{Rasmussen2006}.

\textbf{Factor Graphs}~\cite{kschischang2001factor} are a novel way to represent estimation problems.
A factor graph is a graphical model involving factor nodes and variable nodes.
The variables, or values, represent the unknown random variables in the estimation problem. 
The factors are probabilistic information on the variables and represent measurements, such as of the signal strength, or constraints between values. 
Performing inference in a factor graph is done through optimization, making it suitable for large scale inference problems~\cite{dellaert_factor_2012}.
Factor graphs have been used for very large-scale SLAM problems~\cite{Lamp2} and for estimation of gas concentrations~\cite{gmonroy_time-variant_2016}.

\textbf{Factor Graph based Informative Path planning} allows for information collection in obstacle rich environments.
Factor graphs have been used as a model of a continuous concentration by modeling the problem as a Markov random field. 
Factor graph based informative path planning provides a framework for actively sensing electromagnetic signals which have interactions with the obstacles in the environment~\cite{denniston2023fast}.
This approach has been used to monitor time varying gas distributions in spaces with obstacles by defining a factor graph over the entire space with an unknown value at each cell~\cite{gmonroy_time-variant_2016}. 
It has also been used to jointly estimate the concentration of gas and the wind direction~\cite{gongora_joint_2020}.
%These approaches suffer from a scaling problem due to their ties to the geometry of the environment and do not handle unknown environments.
%TODO cite my other paper

\section{Notation and Formulation}
We use a grid-based representation of the environment for robot movements and measurements.
Let $\gtsensedlocations \subset  \mathbb{R}^2$ be the set of grid positions that the robot can visit and measure the light intensity.
% For a robot that moves in $\mathbb{R}^d$, $\gtlocations \subset \mathbb{R}^d$.
% $\gtsensedlocations$ is the set of locations the robot could measure. 
% If the robot sensor has finer resolution than $\gtlocations$,
% e.g. if the robot uses a camera sensor or takes measurements while traversing between grid points,
% then $|\gtsensedlocations| > |\gtlocations|$.
Let $\vx_t \in \gtsensedlocations$ be the position that the robot visited and measured the light intensity at time step $t$.
Let $\vx_{0:t} = [\vx_0, \ldots, \vx_t]$ be the sequence of visited positions from time $0$ to $t$.
Let $y_t\in \mathbb{R}$ be the measured light intensity at time $t$ at position $\vx_t$, and $y_{0:t} = [y_0, \ldots, y_t]$ be the sequence of measurements.
% $\gtsensedvalues$ and $\sensedvalues_{0:t}$ as the values at all possible measured locations, and the values the robot has measured up to time $t$, respectively.

Configuration $\vs\in \mathbb{R}^{3N}$ of $N$ omni-directional light sources contains position $\vp^i\in\mathbb{R}^2$ and brightness $b^i\in[0,1]$ of each light source $i\in\{1, \ldots, N\}$. 
Let $\vs_t \in \reals^{3N}$ be the configuration of light sources at time step $t$.
% We define $t_{p}$ as the last time step at which the light sources were re-configured.
% Let $T$ be the final step of the operation, and $S_{T}$ is the lighting configuration after the robot has taken $T$ measurements and environment actions.
Let $\obstacles$ be the set of obstacles in the environment, where each obstacle $\obstacles_j \in \obstacles$ is a subset of $\mathbb{R}^2$.
We assume the obstacles are static and known ahead of time.
Let $\mW \subset \mathbb{R}^2$ be the workspace of the operation, which can be set to a rectangular box defining the position bounds, such that $\gtsensedlocations \subset \mW$.

Let $L(\vx, \vs; \obstacles): \gtsensedlocations \times \mathbb{R}^{3N}\rightarrow \mathbb R$ be the light intensity at position $\vx$ when light sources have configuration $\vs$, which is conditioned on the obstacles $\obstacles$ in the environment. 
% We denote the light intensity at time $t$ at each position $x \in\gtsensedlocations$ as $\lightintensity_t(\sensedlocation)=E(\sensedlocation;  \lightlocations_t,\obstacles)$, which is a function of position $x$ conditioned on light source configuration $S_t$ and obstacles $\obstacles$.
Let $\lightintensity^{d}(\vx) : \gtsensedlocations \rightarrow \reals$ be the desired light intensity and $\lightintensity^{u}(\vx) : \gtsensedlocations \rightarrow \reals$ be the unknown light intensity at each position $\vx\in\gtsensedlocations$.\footnote{Note that we do not condition unknown light intensity to obstacles, while in reality, this dependence could exist.}
Let $\lightintensity^s(\vx, \vs; \obstacles): \gtsensedlocations \times \mathbb{R}^{3N}\rightarrow \mathbb R$ be the light intensity at position $\vx$ generated by light sources configured by the robot with configuration $\vs$, conditioned on obstacles.
The light intensity with the light source configuration $\vs$ is given by  $\lightintensity(\vx, \vs; \obstacles) = L^{u}(\vx) + L^{s}(\vx, \vs; \obstacles)$ for each $\vx\in \gtsensedlocations$.

In this work, we aim to find the optimal light source configuration $\vs^{*}$ that minimizes the cumulative difference between the desired light intensities $L^d$ and the actual light intensities $L$ according to the following formulation:
% we need alignat normal because we reference this equation later
\begin{alignat}{2}\label{eq:formulation}
    \vs^* = \argmin_{\vs \in \reals^{3N}} 
 &  \sum_{\vx \in \gtsensedlocations}| \lightintensity^{d}(\vx)  &&- (\lightintensity^{u}(\vx) +  \lightintensity^{s}(\vx, \vs; \obstacles)) | \nonumber \\  
    s.t. \quad  \vp^{i} & \notin  \obstacle_{j} && \forall \obstacles_{j} \in \obstacles,  \forall i \in \{1,\ldots,N\} \nonumber \\
     \vp^i & \in  \mW && \forall i \in \{1, \ldots, N\} \nonumber \\
    % \lightlocations^{n}_{y} & \in  [y_{min}, y_{max}] && \forall n \in N \nonumber\\
     B^{i} & \in  [0,1] && \forall i \in \{1, \ldots, N\}.  
\end{alignat}

\section{Approach}\label{sec:approach}
% \begin{itemize}
%     \item Modeling
%     \begin{itemize}
    
%         \item factor graph
%         \item residual modeling
%     \end{itemize}
%     \item Trigger
%     \begin{itemize}
%         \item discuss logprob
%         \item discuss alpha value
%     \end{itemize}
% \end{itemize}

\begin{figure}[t]
    \centering
    \includegraphics[width=\columnwidth]{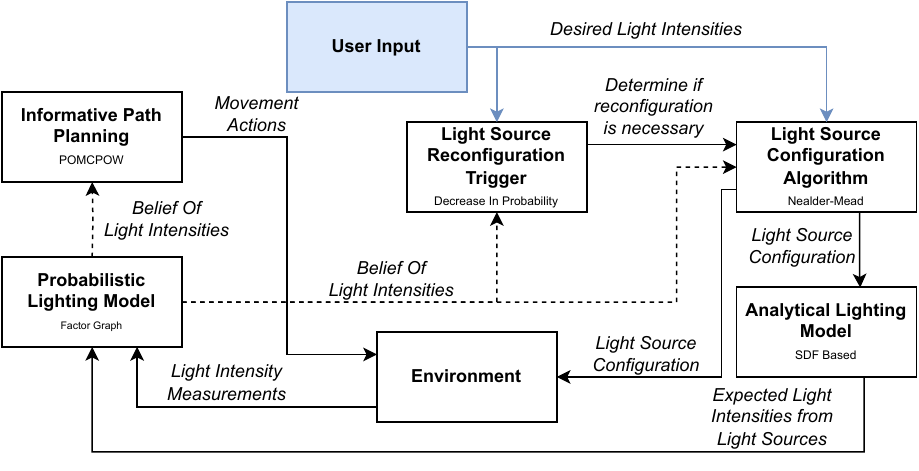}
    \caption{\textbf{Overview of our System:} A description of each module and their connections to other modules are described in \cref{sec:approach}.
    The principle modules we propose in this work are the light source reconfiguration trigger, the light source configuration algorithm and the probabilistic lighting model. 
    }
    \label{fig:system_overview}
    \vspace{-0.25in}
\end{figure}
Here, we describe our system design as well as each module of our system. 
An overview of the system can be seen in~\cref{fig:system_overview}, where we show the inputs and outputs of each module, as well as their connections.

The user specifies the the desired light intensities $L^d(\vx)$ for all $\vx\in \gtsensedlocations$.
The light source reconfiguration trigger determines when the light source configuration algorithm should re-configure the light sources.\footnote{In this work, we do not explicitly model the redeployment process of the light sources. We change positions of light sources immediately in the simulations. A human operator changes light source positions in real world experiments.} 
The analytical lighting model takes this configuration and computes the expected light intensity from the light sources at all positions $\vx \in \gtsensedlocations$. 
The probabilistic lighting model integrates measurements from the environment and the expected light intensities from the analytical lighting model to compute the belief over the light intensities for all $\vx \in \gtsensedlocations$, which is used by the POMDP solver to plan robot movement in the environment to decrease the uncertainty.
At each time step $t$, the environment provides light intensity measurements at the position the robot is currently at, i.e., $\vx_t$, based on both the light source configuration $\vs_t$, the unknown background lighting $\lightintensity^{u}$, and the obstacles $\obstacle$.
We discuss each module in depth in the following sections.

\subsection{Analytical Lighting Model}\label{sec:analytical_lighting}

We use a ray based light propagation model where rays are cast from each light source $i \in \{1, \ldots, N\}$ in a circular pattern until they hit an obstacle and are reflected at the same angle of incidence across the surface normal~\cite{glassner_introduction_1989,distributed_raytracing}.
We use a signed distance function based approach for representing the obstacles in the environment.
We model the light propagation by a ray-marching algorithm~\cite{hart_sphere_1996}.
To determine the intensity of a position $\vx \in \gtsensedlocations$, we use the following equation:
\begin{equation}
    \tilde{L}(\vx) = \sum_{r \in \mathcal{R}_\vx} \left[ \frac{1}{d(\vp_r, \vx)^2} b_r P \prod_{\obstacle_{r}^j \in \obstacle_r}R_r^j \right]
\end{equation}
where $\mR_\vx$ is the set of all rays which pass through $\vx$, $\vp_r$ and $b_r$ are the position and brightness of the light source ray $r$ is originated from respectively, $P$ is the constant scaling factor for light brightness corresponding to the light intensity at maximum brightness $1$, $\obstacle_r$ is the set of all obstacles that ray $r$ reflected off before arriving at $\vx$, $R_r^j \in [0,1]$ is the reflectivity of obstacles $\obstacles_r^j$ and $d(\vp_r, \vx)$ is the distance of the light source generating ray $r$ from $\vx$ along $r$.

While this lighting model is considerably simpler than modern photo realistic lighting models~\cite{parker_gpu_2013}, we find that it provides a reasonable approximation of the real lighting conditions when we compare predicted light intensities with collected sensor measurements in field trials in \cref{sec:field_experiments}, especially when considering the noise present in commodity lighting sensors.

\subsection{Probabilistic Lighting Model}\label{sec:light_belief}
The model described in \cref{sec:analytical_lighting} allows the robot to determine the expected lighting based on where the light sources are placed, but the model ignores many components of actual light propagation such as refraction. 
Additionally, the model has no way of incorporating previously unknown background lighting. 
For this reason, analytical lighting model alone is often inaccurate as we show in our experimental results. 

In order to allow the robot to both integrate sensor measurements into the analytical predictions and determine the uncertainty of predicted light intensities, we use a probabilistic estimation model.
% The first we term a residual model. 
The model is inspired by residual physics models~\cite{residual_tossing} %ask david 
in which a simulated physical model is corrected with a learned model.
This, for example, can use a traditional Gaussian process to model the difference in the output of the analytical lighting model and the measured light intensity.
However, using Gaussian processes in our case presents two major problems:
i) Gaussian processes have difficulty representing obstacles and ii) Gaussian processes cannot represent the uncertainty due to both the analytical lighting model's inability to model all the effects present in real light propogation as well as the noise on the sensor measurements, as a Gaussian process can only represent this as one combined uncertainty.
\begin{figure}
    \centering
    \includegraphics[width=\columnwidth]{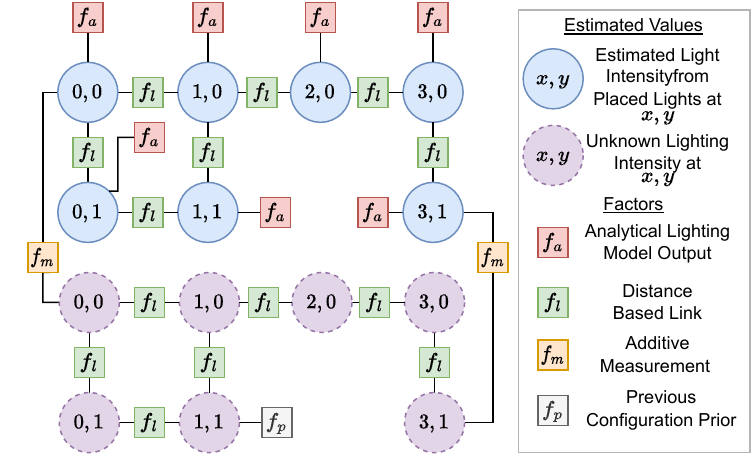}
    \caption{\textbf{Probabilistic Lighting Model.} This model, described in \cref{sec:light_belief}, allows the robot to combine the ray marching based analytical lighting model for the configured light sources with the real-world measurements of the light intensities using additive measurement factors, $f_m$. In this example, the robot has taken a measurement at locations $(0,0)$ and $(3,1)$. The robot has previously taken a measurement at $(1,1)$ before the light sources were re-configured, which is reflected to the previous configuration factor, $f_p$. There is an obstacle at $(2,1)$, therefore, there are no variables there and distance based link factors, $f_l$, are removed.}
    \label{fig:fg}
    \vspace{-0.2in}
\end{figure}
To tackle these issues, we propose a factor graph based model which incorporates four types of factors and two types of variables, as can be seen in \cref{fig:fg}.

The first type of variables ({\color{blue}blue} variables in \cref{fig:fg}) represent the estimated true light intensity generated by the configured light sources at each position $\vx\in \gtsensedlocations$, corresponding to $L^s(\vx, \vs; \obstacles)$, which are connected to unary measurement factors ($f_{a}$ in \cref{fig:fg}) based on the output from the analytical lighting model.
The second type of variables estimate the unknown light intensity at each position $\vx\in\gtsensedlocations$ ({\color{violet}purple} variables in \cref{fig:fg}), corresponding to $L^u(\vx)$.
Both types of variables are linked to their neighbors by a distance based link factor ($f_{l}$ in \cref{fig:fg}) when there is not an obstacle in between.
This allows the factor graph to model the estimation problem around the topology of the environment.
The uncertainty in these factors is based on the spatial distance between the values, similar to the length-scale in a Gaussian Process.

In order to integrate sensor measurements into the system, we propose a novel factor called the additive measurement factor ($f_{m}$ in \cref{fig:fg}).
At each time step $t$, the sensor reads the light intensity $y_t$ at the current position $\vx_t$.
The robot does not know how much of the intensity comes from the configured light sources and how much is unknown light.
This factor models the measurements the sensor has taken as being $y_t = \lightintensity^{u}(\vx_t) + \lightintensity^s(\vx_t, \vs_t; \obstacles)$.
To achieve this, we add a factor with a likelihood $\mathcal{N}(v^{s}(\vx_{t}) + v^{u}(\vx_{t}) | y_t, \sigma^2)$, where $v^{p}(\vx_{t})$ and $v^{u}(\vx_{t})$ are the estimated light intensity from configured light sources and estimated unknown light intensity at $\vx_t$ respectively.

When the robot re-configures the lights, the additive measurement factor is no longer valid as $\lightintensity^{s}$ changes.
To allow the robot to continue using its previous measurements, we add a unary factor called the previous configuration factor ($f_{p}$ in \cref{fig:fg}) to each unknown lighting intensity variable where an additive measurement factor has previously been. 
The mean and variance of this factor are the posterior mean and variance for this unknown lighting variable.

Solving the factor graph amounts to running an optimizer to find the maximum a-posteriori (MAP) values for the values, e.g. the most likely light intensity at each location.
Finding the uncertainty of the estimated light intensity for each location requires computing the Gaussian marginals for each value~\cite{dellaert_factor_2012}.

The output of the probabilistic lighting model is a set of normal distributions with mean $\mu_t(\vx)$ and standard deviation $\sigma_t(\vx)$ for each $\vx \in \gtsensedlocations$ at each time step  $t$.

\subsection{Informative Path Planning}
We treat the problem of determining the optimal path to move the robot through the environment to collect light intensity measurements as an IPP problem which we solve with a rollout based POMDP solver, based on 
the POMCPOW method~\cite{sunberg_online_2018,denniston_icra_2021}.
In informative path planning, the robot iteratively takes actions and measurements to maximize an information-theoretic objective function~\cite{Hollinger2014}.
The IPP objective function we use is the entropy of each $\vx \in \gtsensedlocations$ similarly to ~\cite{guestrin2005,low_information-theoretic_2013}, which relates to how well the model spatially covers the workspace:
\begin{equation}\label{eq:entropy}
   f(\vx) = \frac{1}{2} \ln(2 \pi e \sigma_t^2(\vx)).
\end{equation}
where $t$ is the time step planning has started.
This allows the robot to iteratively decrease the uncertainty in the lighting intensity by maximally reducing the uncertainty in the model along a trajectory.

We adopt the t-test heuristic proposed by Salhotra et. al~\cite{icra_paper} for taking multiple actions from one plan.

\subsection{Light Source Reconfiguration Trigger}\label{sec:trigger}
To determine when the robot should reconfigure the light sources, we propose a likelihood based trigger. 
This trigger allows the robot to compute a new configuration when necessary, creating a balance between reconfiguring the light sources in accordance with the new information, while simultaneously allowing the robot to have enough measurements with each light configuration so that it can deduce where the light intensities are different than the desired ones.

To this end, we propose a trigger which looks at the likelihood of observing the desired light intensities ($\lightintensity^{d}$) given the current estimated light intensities obtained from the probabilistic lighting model at each time step $t$:
\begin{equation}
    \mathcal{L}_t(L^{d}) = \prod_{\vx \in \gtsensedlocations} \mathcal{N}\left(L^{d}(\vx) \mid \mu_t(\vx), \sigma_t(\vx) \right). 
\end{equation}
We assume that the observation likelihoods across grid positions are independent, while in reality, this dependence exists. 
The independence assumption allows us to compute the joint observation likelihood in a computationally efficient way.

In order to determine whether to reconfigure the light sources, the robot determines if the likelihood of observing the desired lighting intensity has fallen signifigantly since the last time step $t_c$ at which light sources are reconfigured. 
This occurs when the robot collects measurements that make observing the desired lighting less likely given the current configuration.
To this end, the robot re-configures the lights when the following inequality holds
\begin{equation}
    log \left[ \mathcal{L}_t(\lightintensity^{d}) \right] < \alpha\max_{\tau \in [t_{c}:t-1]} log \left[ \mathcal{L}_\tau(\lightintensity^{d}) \right]\label{eqn:trigger}
\end{equation}
where $\alpha \in [1, \infty)$ is an experimentally chosen scaling constant.
Once the log-likelihood has sufficiently decreased from it's maximal since last reconfiguration, the robot re-configures the light sources and calculates a new configuration.

\subsection{Light Source Configuration Algorithm}
We solve the optimization problem described in \cref{eq:formulation} to reconfigure the light sources, which includes computing position $\vp^i$ and brightness $b^i$ of each light source $i\in\{1, \ldots, N\}$, which are the decision variables of our optimization formulation.
% In order to compute a new light source configuration to achieve the minimal error described in \cref{eq:formulation}, we must re-optimize the locations and brightnesses of the placed lights once re-configuration is triggered according to \cref{sec:trigger}. 
Because $\lightintensity^{u}$ and $\lightintensity^{s}$ cannot be measured directly, we use the estimates of $\lightintensity^u$ from the probabilistic lighting model, and use analytical lighting model to predict $\lightintensity^s$.
We minimize the error to find the optimal lighting configuration given the incomplete information in the probabilistic lighting model and inaccuracies of the analytical model.

We use the Nelder-Mead optimizer to minimize this quantity~\cite{olsson_nelder-mead_1975}.

\section{Experiments}
We demonstrate the effectiveness of the components of our proposed system when compared to alternatives as well as the overall effectiveness of it when compared to a baseline.
We also demonstrate our system in a real field test scenario, in which light sources are placed in a warehouse environment.

In all experiments, we provide the user input, i.e., the desired light intensities $L^d$, by randomly configuring light sources in an obstacle-free environment, and computing analytical light intensities from the configured light sources, which the robot must match in an environment with obstacles.
The robot is given a fixed number of light sources to configure in order to match the desired light intensities generated by the randomly configured light sources. 
The robot always configures all available light sources, i.e, it does not configure them sequentially.
We do not explicitly model the redeployment process of the light sources and reflect new configurations to the environment immediately.
In real world experiments, a human operator changes the light source positions before robot continues operation.
In simulations, we create unknown light sources at half brightness randomly placed in the environment.
We use 1 reflection while predicting $L^s$ using analytical lighting model, but use 5 reflections while providing measurements to the robot. This is done to increase the error in the analytical lighting model used by the robot, simulating unmodeled effects in the analytical lighting model.
In simulations, the robot is able to control the brightness of the lights to match the desired light intensities, while in real-world experiments, the brightness of light sources are fixed and the robot is able control the positions of light sources only.
The robot operates in a $13\times 13$ grid environment with a grid spacing of $0.35m$.
The robot can take four actions (up, down, left, right) at each grid cell (unless it would run into an obstacle).
Each experiment begins with the same initial light source configuration for the same environment number and seed.

The robot is allowed to take $100$ environment steps in simulation experiments.
Each step consists of a movement action followed by a sensing action once the robot arrives at the desired location, possibly followed by light configuration depending on the light source reconfiguration trigger.
The robot is allowed to take $70$ environment steps in the field trial due to time constraints. 
The planner is given $50$ rollouts to evaluate future actions. 
% For all simulation experiments, we plot the RMSE for each environment and seed.
We differentiate between environments (seeds for the placement of obstacles, ambient lights and desired lights) and seeds each (seeds for the starting location of the robot) to allow for diverse environment generation.

\subsection{Component Experiments}
We present experiments which demonstrate the improvement of our system due to the proposed components.
For each component test, the robot places $10$ light sources to match desired light intensities of $7$ randomly placed lights. 
We place $40$ unknown light sources to create randomly to create unknown light intensity $L^u$.

\subsubsection{Probabilistic Lighting Model}

We show the effectiveness of our factor graph based probabilistic lighting model proposed in \cref{sec:light_belief}, by comparing it with a residual Gaussian process.
The residual Gaussian process models the difference between the analytical lighting model and the real world measurements.
The residual Gaussian process uses a squared exponential kernel, which is standard in informative path planning~\cite{kemna_voronoi}.
The residual Gaussian process can provide uncertainties and predictions at each position $\vx \in \gtsensedlocations$.

We configure the light sources every $10$ environment steps to provide comparison. 

\begin{figure}
    \centering
    \includegraphics[width=\columnwidth]{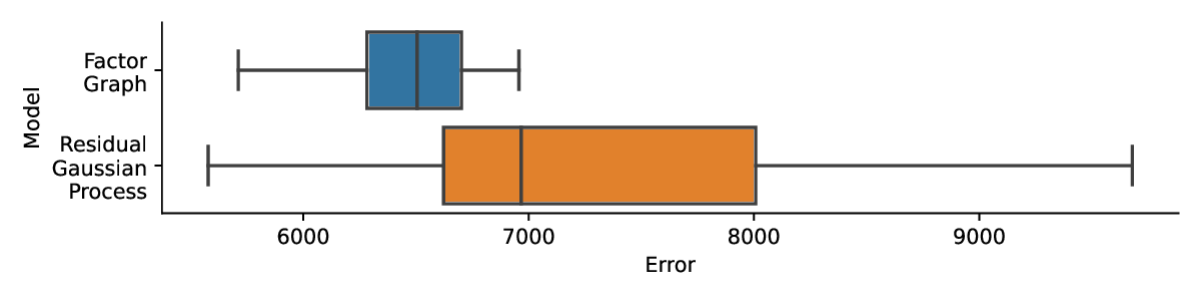}
    \caption{\textbf{Light Belief Model Results} We show the reduction in error when using the model proposed in \cref{sec:light_belief} when compared to a Gaussian process baseline model.}
    \label{fig:model_results}
    \vspace{-0.2in}
\end{figure}
We compute the average RMSE over each measured location across five environments with five seeds each, totaling $25$ trials for each model, shown in~\cref{fig:model_results}.
Our model provides a large reduction in final RMSE.
We believe this is due to the ability of our model to represent i) obstacles, ii) errors in the analytical model  better than the baseline model.
The factor graph based model produces qualitatively better trajectories which take obstacles into account during informative path planning.
This can be seen in~\cref{fig:hero_setup} where the proposed trajectories cover the space well.

\subsubsection{Light Source Reconfiguration Trigger}
% The light source reconfiguration trigger determines when the robot should configure the light sources.
% When the light source reconfiguration trigger is determines that the light sources should be reconfigure, the robot runs an optimizer to find the new optimal configuration of the light sources given its prediction about the unknown light intensities and effects, and the newly configured lights are deployed.
We compare our proposed trigger in~\cref{sec:trigger} against three baselines.
The first baseline does not use any of the collected measurements after the initial light source configuration and simply uses the analytical model. 
We term this trigger \textit{First Step}, which represents configuring light sources using a priori information.
The second baseline uses the initial configuration of light sources for the entire measurement mission and after the very last step, reconfigures the lights.
We term this \textit{Last Step}, which represents conducting a sampling mission without interleaving light re-configuring actions.
The third baseline re-configures the light sources at set intervals of environment steps, which we call the \textit{every $n$} baseline.
For example, \textit{Every $10$} re-configures the light sources every $10$ steps. 
We consider this the strongest trigger baseline as it provides a parameter ($n$) which can be optimized to provide a strong behavior.
We compare these baselines against our proposed trigger, given in \cref{eqn:trigger}, which we call the \textit{Logprob} trigger.
This trigger looks at the decrease in log-likelihood of observing the desired light intensity given the current model, and re-configures the light sources once it is below $\alpha$ times the last highest log-likelihood. 

\begin{figure}
    \centering
    \includegraphics[width=\columnwidth]{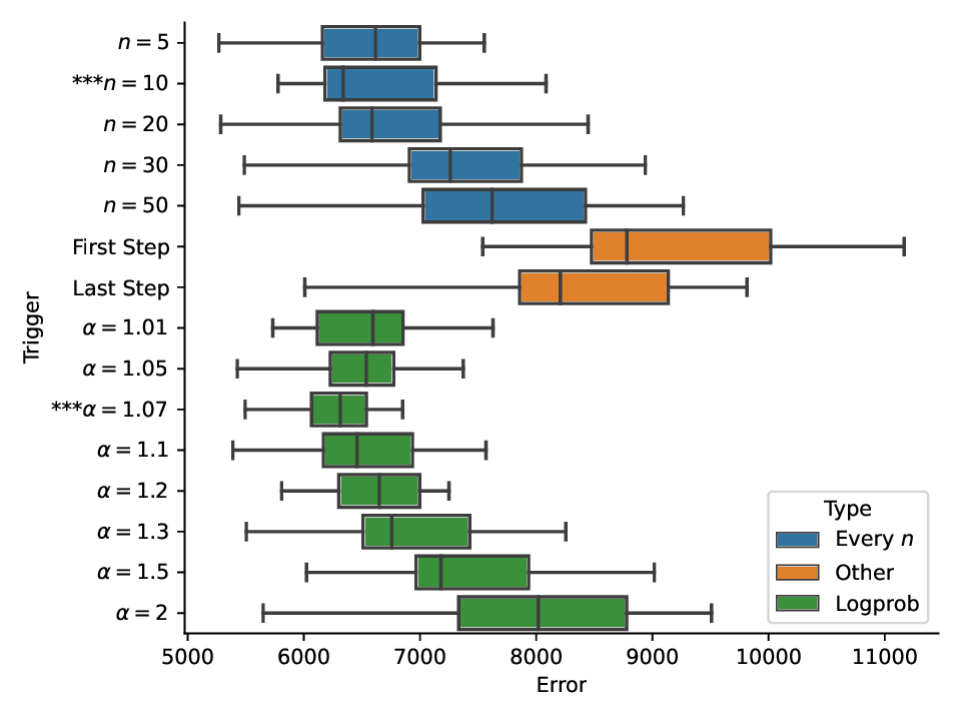}
    \caption{\textbf{Light Source Reconfiguration Triggers.} Comparison of RMSE when different triggers with different parameters are used. We see that there is a balance between reconfiguring the light sources more and less often in both \textit{Logprob} and \textit{Every $n$} triggers. \textit{Logprob $\alpha=1.07$} results in lowest RMSE.}
    \label{fig:trigger_results}
\end{figure}
\begin{figure}
    \centering
    \includegraphics[width=\columnwidth]{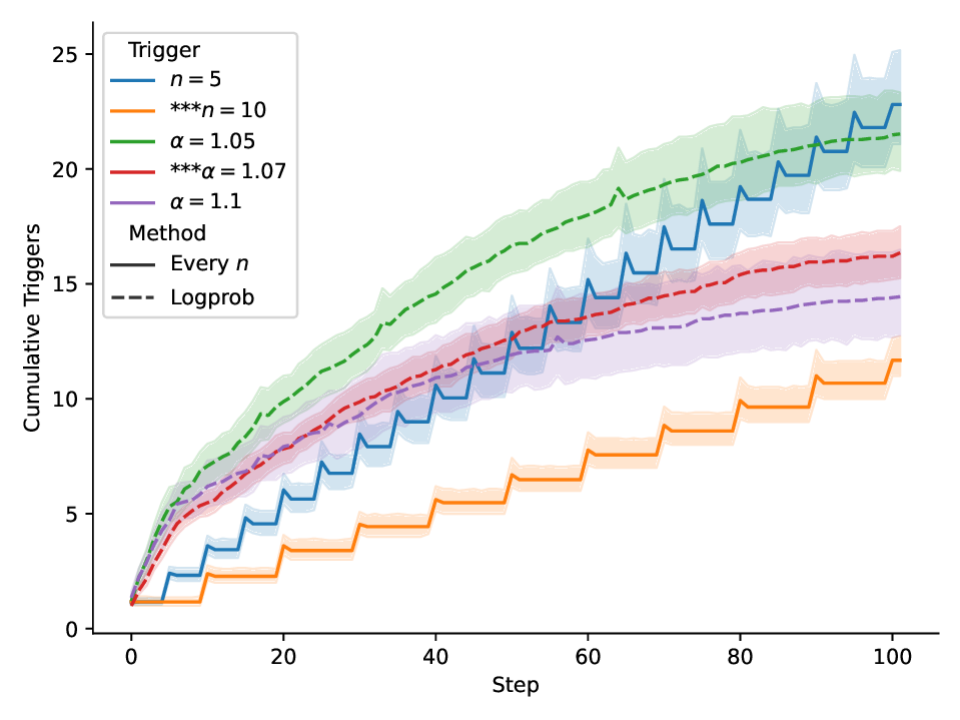}
\caption{\textbf{Cumulative Number of Light Sources Reconfigurations.} The cumulative number of times the robot has re-configured the lights over steps in the environment. The \textit{Logprob} triggers tend to use more triggers early in the task when the uncertainty about unknown light sources is high.}
    \label{fig:cumulative_triggers}
    \vspace{-0.2in}
\end{figure}

We show the results for various configurations of triggers across five environments with three seeds each in \cref{fig:trigger_results}. 
We find that \textit{First Step} and \textit{Last Step} are both outperformed by all of the attempted \textit{Every $n$} and \textit{Logprob} triggers.

For the \textit{Every $n$} triggers, we find that $n=10$ is optimal.
We see that there is a balance between re-configuring the lights too frequently and too infrequently, where 10 appears to strike a balance in our experiments.

We see a similar trend of balancing how often the lights are re-configured with the \textit{Logprob} triggers.
We see that $\alpha=1.07$ is optimal, with a lower median and upper quantile error than optimal \textit{Every $n$} trigger.

To show how the number of light source re-configurations increase over the task, we plot the cumulative number of light source reconfiguration triggers over environment steps in~\cref{fig:cumulative_triggers}.
We see that \textit{Every $n$} and \textit{Logprob $\alpha=1.07$} converge to the same amount of cumulative triggers, but differ in where they are used.
\textit{Logprob $\alpha=1.07$} tends to result in more light source reconfigurations at the beginning when the robot is rapidly decreasing its uncertainty in the unknown light distribution, and tends to decrease the number of light source reconfigurations at the end.
This shows that not only the number of triggers, but temporal distribution matters for this task.

\subsection{System Experiments}\label{sec:system_exeperiments}
\begin{table}
\caption{Task complexity levels for system experiments. The number of obstacles, number of unknown light sources, number of target light sources used to generate the desired light intensities, and number of configured light sources that the robot can control increase from task level A to C.}
\begin{tabular}{ m{0.1\columnwidth} || m{0.14\columnwidth} m{0.14\columnwidth} m{0.14\columnwidth} m{0.14\columnwidth}}
 % \multicolumn{5}{c}{Task Complexities} \\
 % \hline
 Level & Obstacles & Unknown Light S. & Target Light S. & Configured Light S. \\ 
 \hline \hline
 A & 8 & 40 & 7 & 10 \\
 B & 12 & 60 & 9 & 15 \\
 C & 16 & 70 & 12 & 20 \\
\end{tabular}
\label{tbl:tasks}
\vspace{-0.1in}
\end{table}

\begin{figure}
    \centering
    \begin{subfigure}{\columnwidth}
    \includegraphics[width=\textwidth]{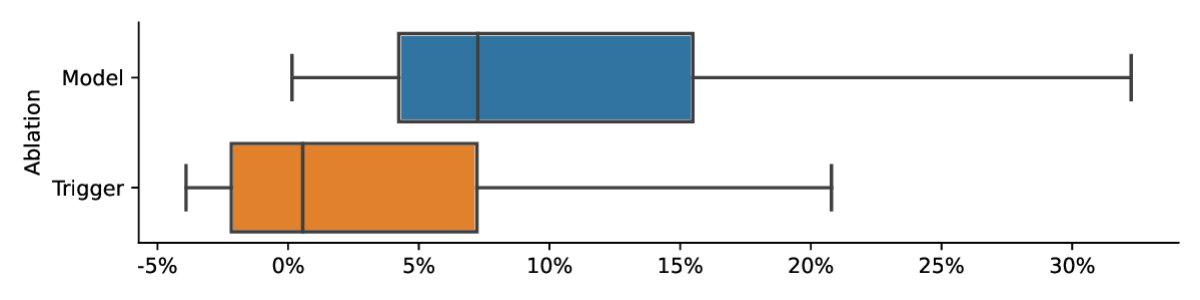}
    \end{subfigure}
    \begin{subfigure}{\columnwidth}
    \includegraphics[width=\textwidth]{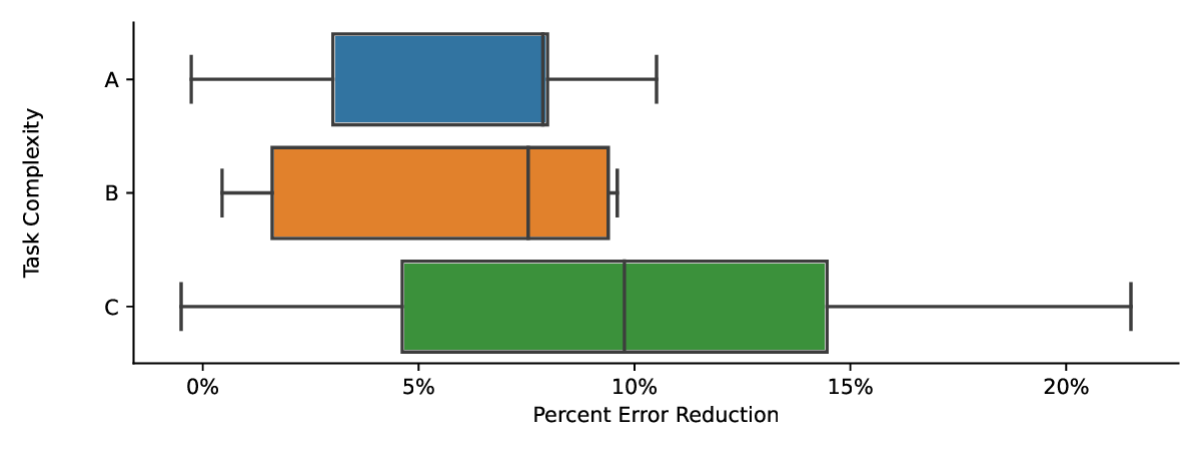}
    \end{subfigure}
    \caption{\textbf{Module and System Improvement:} Top: Improvement in our system due to various per-module improvements vs baseline modules, Bottom: Improvement in our system across various environment difficulties, see \cref{sec:system_exeperiments} for a description of the task complexities. We report percent error due to differences in error magnitude between complexities and seeds.}
    \label{fig:improvements}
    \vspace{-0.25in}
\end{figure}

% \subsubsection{baseline}
% we compare our proposed system with teh factor graph model 
% compare in different scenarios against planning every step and only planning after ipp 
We compare our system to a baseline system across various task complexities.
The baseline system consists of the residual Gaussian process model and the \textit{Every $10$} trigger.
The proposed system uses our novel factor graph based probabilistic lighting model proposed in \cref{sec:light_belief} and the \textit{Logprob $\alpha=1.07$} trigger proposed in \cref{sec:trigger}.

We compare across three task complexities described in~\cref{tbl:tasks}. 
Task complexity A is the same task complexity used in the component experiments.
Level B is harder than level A, and level C is harder than level B.
For each of these task complexities, we use five different environments (seeds for the placement of obstacles, ambient lights and desired lights) with five seeds each (seeds for the starting location of the robot).

The results are given in \cref{fig:improvements} as improvement percentages of the proposed system over the baseline system in final RMSE.
The proposed system provides a median improvement of $7.9\%$, $7.5\%$, and $9.8\%$ in RMSE on task complexities A,  B, and C respectively.
This demonstrates that, our system considerably reduces the error, especially in difficult tasks. 

We also see in \cref{fig:improvements} that the factor graph based probabilistic lighting model provides a larger improvement than the light source reconfiguration trigger, implying that modeling the obstacles provides a greater benefit than accurately choosing when to re-configure the lights.

 \subsection{Field Tests}\label{sec:field_experiments}
\begin{figure}
    \centering
    \begin{subfigure}{0.45\columnwidth}
    \includegraphics[width=\textwidth]{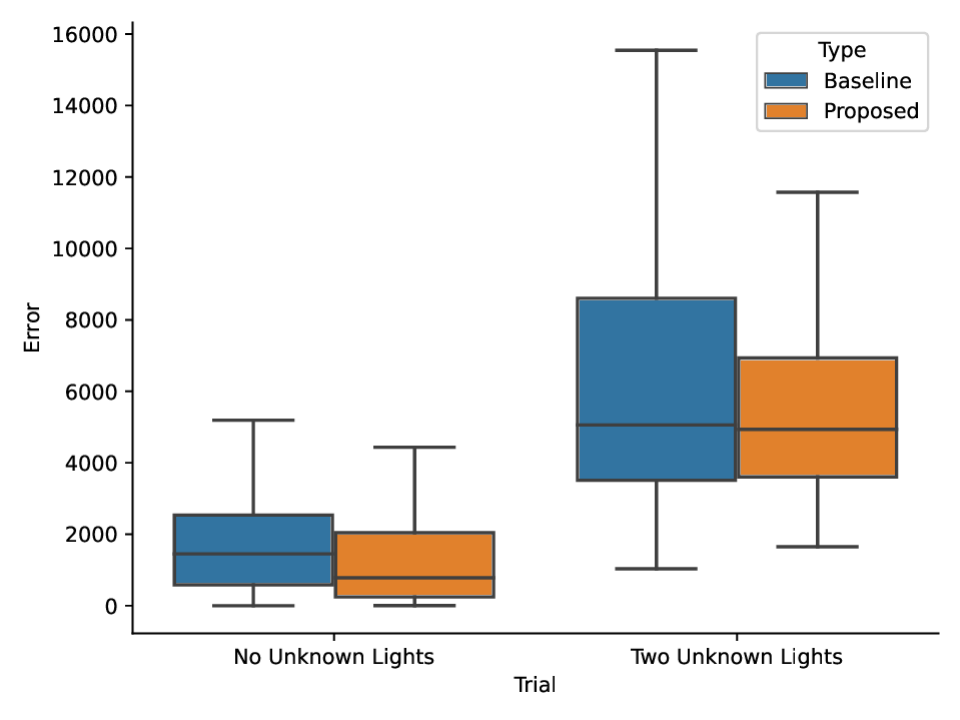}
    \caption{Point based RMSE}
    \end{subfigure}
    \begin{subfigure}{0.45\columnwidth}
    \includegraphics[width=\textwidth]{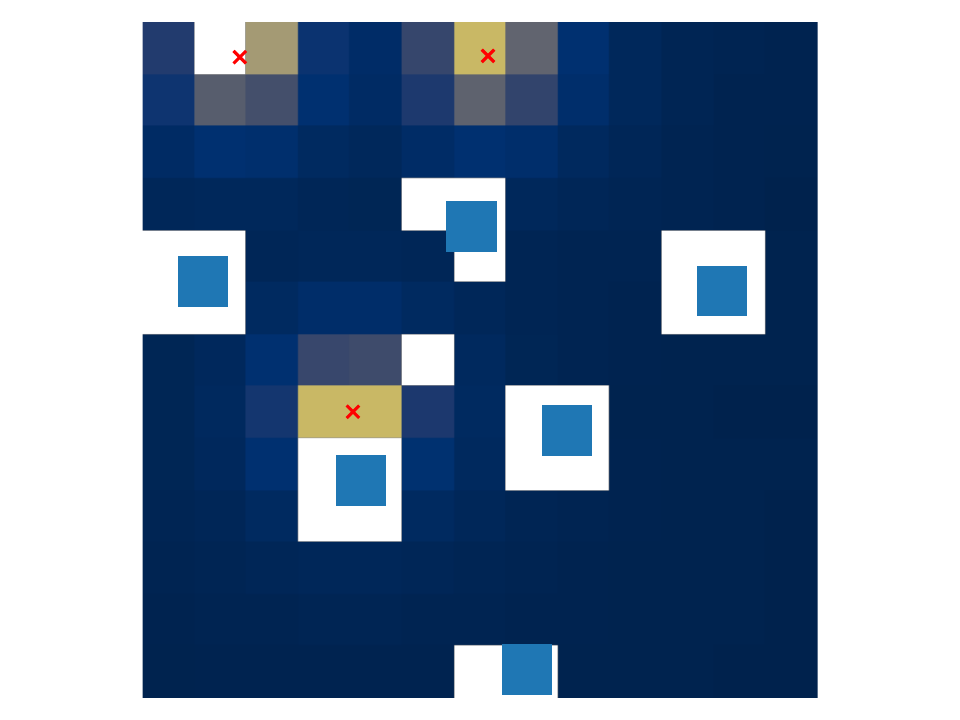}
    \caption{Desired}
    \end{subfigure} \\
    
    \begin{subfigure}{0.45\columnwidth}
       \includegraphics[width=\textwidth]{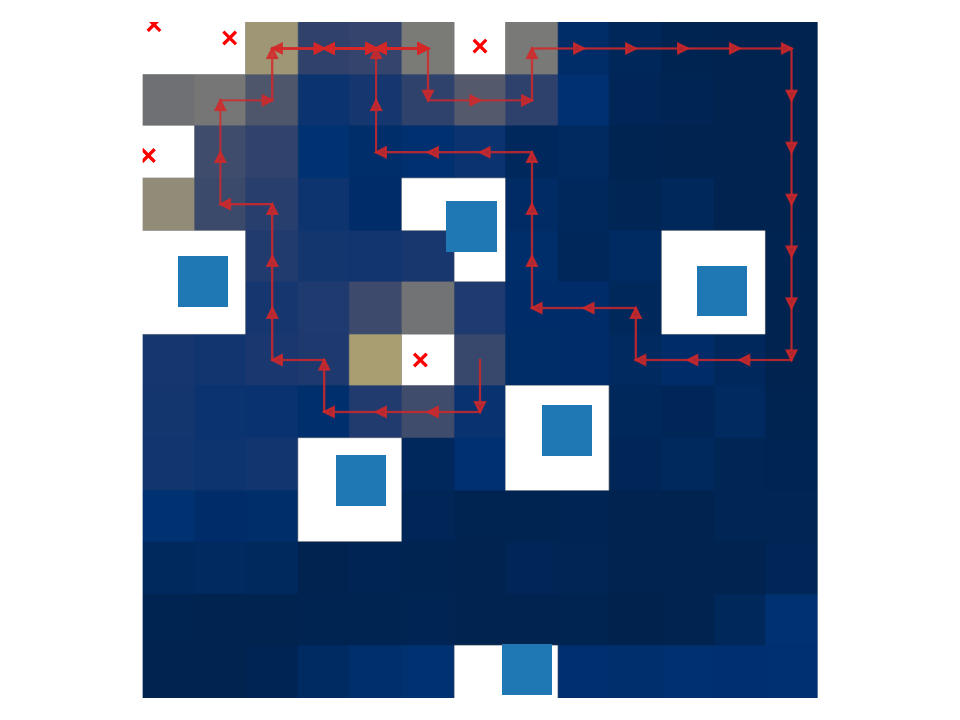} 
       \caption{Proposed: No Unknown}
    \end{subfigure}
    \begin{subfigure}{0.45\columnwidth}
       \includegraphics[width=\textwidth]{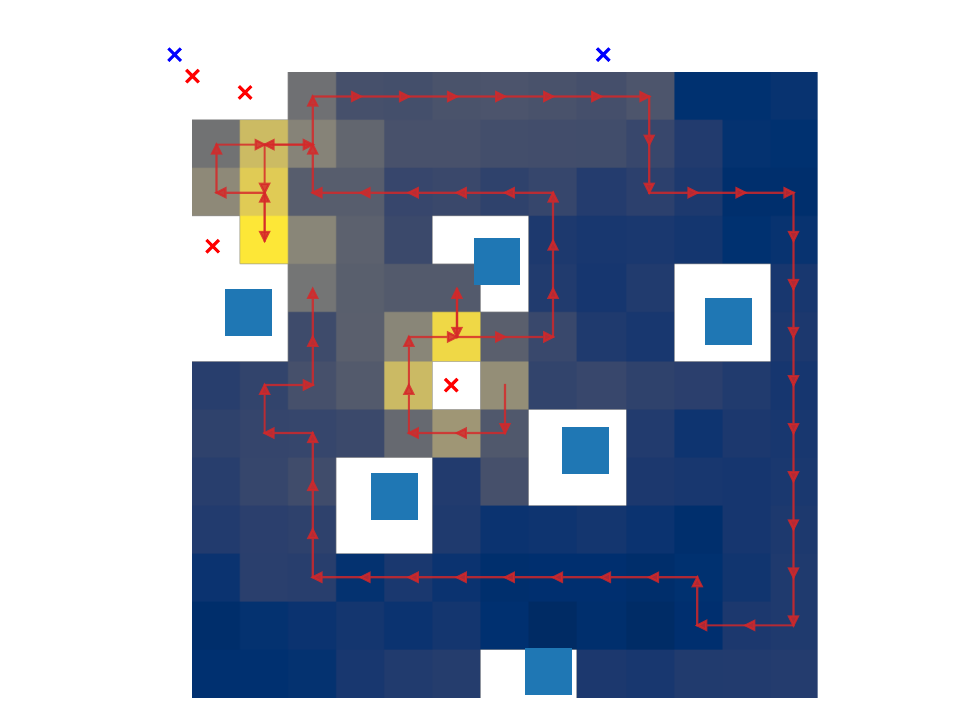} 
       \caption{Proposed: 2 Unknown}
    \end{subfigure} \\
    %\vspace{-.2cm} 
    
    \begin{subfigure}{0.45\columnwidth}
    \includegraphics[width=\textwidth]{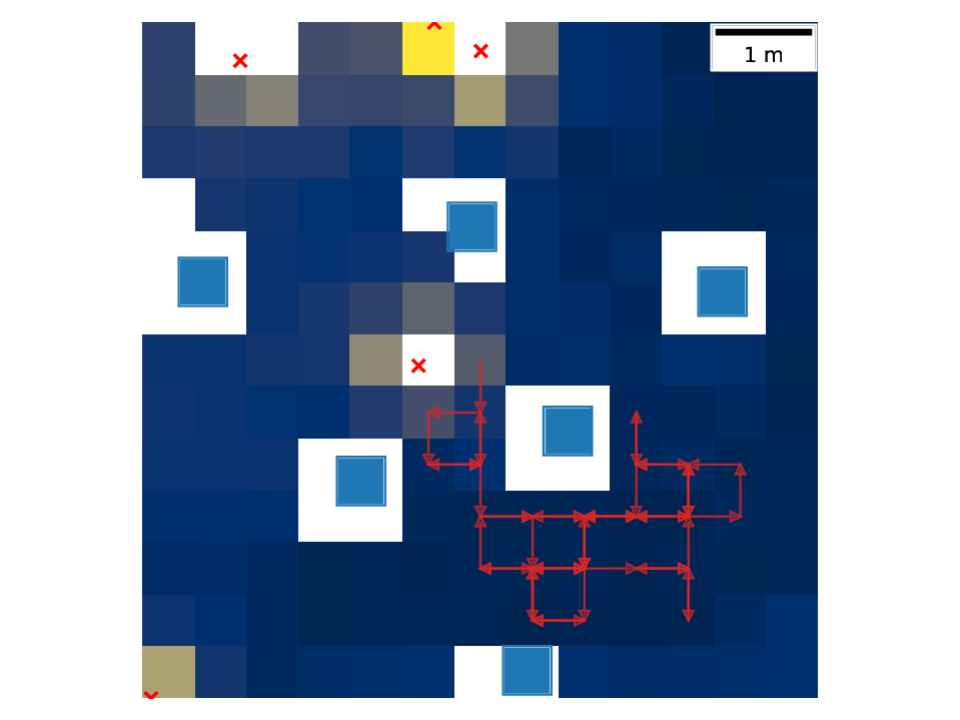}
    \caption{Baseline: No Unknown}
    \end{subfigure}
    \begin{subfigure}{0.45\columnwidth}
    \includegraphics[width=\textwidth]{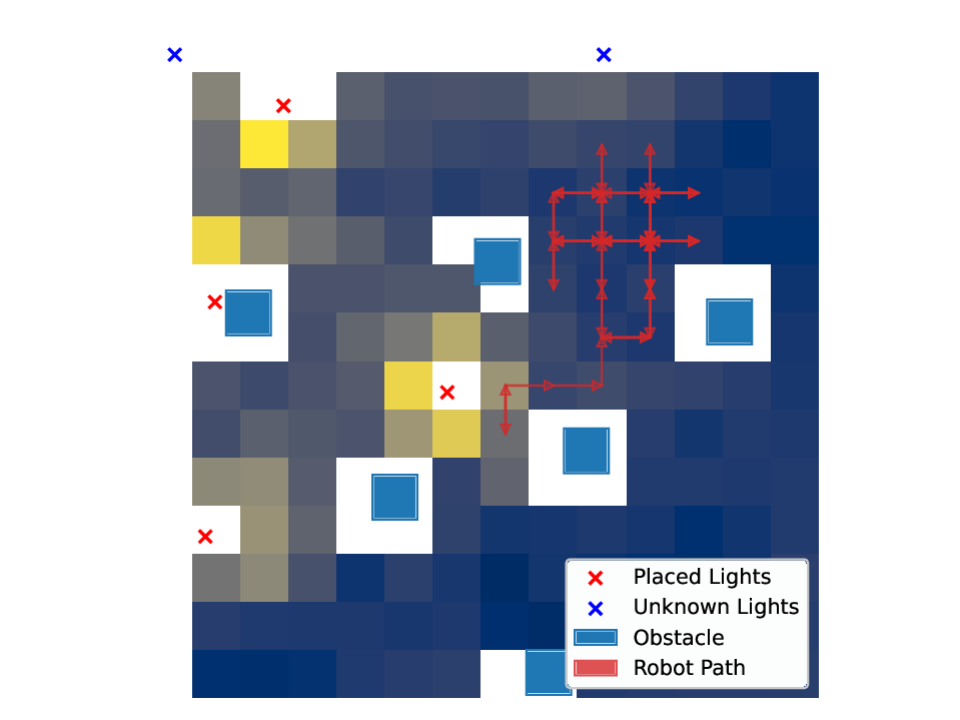}
    \caption{Baseline: 2 Unknown}
    \end{subfigure}
    \caption{\textbf{Field Test Results} We compare our approach in a scenario with a real robot and light sources. We find that our approach is better able to find light source configurations which match the desired lighting configuration in the field. Background color represents post-experiment measured light brightness.}\label{fig:field_results}
    \vspace{-0.2in}
    \end{figure}
We conduct a test of our system on a Turtlebot2 robot with a light sensor.
We test in two types of environments: one with no purposely placed unknown light sources, and one with two purposely placed unknown light sources.\footnote{We call these unknown light sources purposely placed, because there are other light sources in the environment, e.g., lights of computers in the environment as well as light reflected from the ceiling of the experiment space, that are not purposely placed.}
These tests are conducted in a warehouse with light bleed due to windows causing unknown lighting in both environments.
Each experiment has desired light intensities generated from three light sources.
There are five obstacles, which are $0.5\times0.5\ m^2$ cardboard boxes. 
Each environment uses the same desired lighting configuration and obstacle configuration.
In the environment with no unknown light sources, the robot configures five light sources, and in the environment with two unknown light sources the robot places four light sources.
The configured light sources are 1440 lumen battery powered lights placed high enough for the sensor on the robot to detect.
The unknown light sources are 1800 lumen lights plugged into wall power. 
The environment can be seen in \cref{fig:field_hero}.\looseness=-1

After the robot has taken $70$ actions, the robot goes to each $\vx \in \gtsensedlocations$ and collects measurements to determine how accurately the final lighting intensities compared to the desired lighting intensities.

We show the results in \cref{fig:field_results}, where the per-point error is computed for each environment.
We see that our approach provides a considerably lower error in the environment with no purposely added unknown light sources, which is the nominal operating regime for this kind of system.
We also see that when we add two bright unknown light sources, our approach is able to achieve a lower upper quartile error and lower maximum error, but comparable median error. 
We believe this is because this environment is very complicated and the high upper quantile error from the baseline is due to not incorporating the light from the unknown light sources.

\section{Conclusion}
Autonomously deploying light sources and other electromagnetic signal emitters is an important and challenging task for indoor navigation, localization, and connectivity of humans and robots.
We have shown that by interleaving light placement and active sensing, the accuracy of electromagnetic signal emitter placement can be increased.
We also have shown that integrating an analytical model of the light propagation from placed light source into the belief modeling allows for more accurate modeling of both the light intensities in the environment, and the uncertainty associated with them.
We have also shown that this model can be used to decide when to reconfigure light sources, proposing a custom trigger using the likelihood of the desired lighting being measured.
We demonstrate the effectiveness of our system on both simulated and real-world experiments, showing that our method outperforms a baseline system.
The presented approach can readily be extended to other electromagnetic signals as long as an analytical model for them is provided and the robot has sensors to measure the signal strength.

% \bibliographystyle{IEEEtran}
% \bibliography{references}
\printbibliography
\end{document}